\pdfoutput=1
\documentclass[conference]{IEEEtran}
\IEEEoverridecommandlockouts
\usepackage{cite}
\usepackage{amsmath,amssymb,amsfonts}
\usepackage{algorithmic}
\usepackage{graphicx}
\usepackage{textcomp}
\usepackage{xcolor}
\def\BibTeX{{\rm B\kern-.05em{\sc i\kern-.025em b}\kern-.08em
    T\kern-.1667em\lower.7ex\hbox{E}\kern-.125emX}}
\usepackage{algorithm}

\usepackage{array}
\usepackage{subfigure}
\usepackage{multirow}   
\usepackage[normalem]{ulem}
\usepackage{stmaryrd}
\usepackage{diagbox}
\usepackage{comment}
\usepackage{url}
\usepackage{slashbox}
\usepackage{pdfpages}
\usepackage{caption}

\graphicspath{{./image/}}
\usepackage{hyperref}
\usepackage{color}
\usepackage[normalem]{ulem}
\newcounter{ToDo}
\newcounter{gaocomm}
\newcounter{Note}
\definecolor{blue-violet}{rgb}{0.54, 0.17, 0.89}
\definecolor{mygreen}{rgb}{0.0, 0.5, 0.0}
\definecolor{awesome}{rgb}{1.0, 0.13, 0.32}
\definecolor{bostonuniversityred}{rgb}{0.8, 0.0, 0.0}

\newcommand{\GaoC}[1]{\textcolor{blue-violet}{\stepcounter{gaocomm}{\bf [Junbin's comment \arabic{gaocomm}: #1]}\;}}

\newcommand\myeq{\mathrel{\stackrel{\makebox[0pt]{\mbox{\tiny def}}}{=}}}
\begin{document}
\title{Sparse Least Squares Low Rank Kernel Machines\thanks{Both Di Xu and Manjing Fang are students who enroll in Master of Commerce at the University of Sydney and have equal contribution to the project.}}
\author{\IEEEauthorblockN{Di Xu and Manjing Fang}
\IEEEauthorblockA{\textit{Discipline of Business Analytics}\\
\textit{The University of Sydney Business School} \\
\textit{The University of Sydney}\\
Sydney, NSW 2006, Australia \\
\{dixu3140,mfan9400\}@uni.sydney.edu.au} 
\and 
\IEEEauthorblockN{Xia Hong}
\IEEEauthorblockA{\textit{Department of Computer Science} \\
\textit{University of Reading}\\
Reading RG6 6AH, UK\\
x.hong@reading.ac.uk}
\and
\IEEEauthorblockN{Junbin Gao}
\IEEEauthorblockA{\textit{Discipline of Business Analytics}\\
\textit{The University of Sydney Business School} \\
\textit{The University of Sydney}\\
Sydney, NSW 2006, Australia \\
junbin.gao@sydney.edu.au}
}


\maketitle

\begin{abstract}
A general framework of least squares support vector machine with low rank kernels, referred to as LR-LSSVM, is introduced in this paper. 
The special structure of low rank kernels with a controlled model size brings sparsity as well as computational efficiency to the proposed model. 
Meanwhile, a two-step optimization algorithm with three different criteria is proposed and various experiments are carried out using the example of the so-call robust RBF kernel to validate the model. The experiment results show that the performance of the proposed algorithm is comparable or superior to several existing kernel machines.
\end{abstract} 


\section{Introduction}



With the proliferation of big data in scientific and business research, in practical nonlinear modeling approaches, one wishes to build sparse models with more efficient algorithms. Kernel machines (KMs) have attracted great attention since the support vector machines (SVM), a well linear binary classification model under the principle of risk minimization, was introduced in earlier 1990s \cite{SchlkopfSmola2002}. In fact, KMs have extended SVM by implementing the linearity in the so-called high dimensional feature space under a feature mapping implicitly determined by a Mercer kernel function. 
Both SVM and KMs have been also applied for regression problems \cite{Bishop2006}. Commonly used kernels are radial basis function kernel (RBF), polynomial kernel, and Fisher kernel \cite{jaakkola_haussler_1998}, 
etc. As one of the most well-known members of the KM family, SVM has the advantages of good generalization and insensitivity to overfitting \cite{lotte_congedo_lecuyer_lamarche_arnaldi_2007}. 

Until now  Gaussian RBF kernel is the most common choice for SVM in practice. Generally, SVM with RBF kernel has been widely used and has superior prediction performance in many areas such as text categorization \cite{joachims_1998}, image recognition \cite{gumus_kilic_sertbas_ucan_2010}, bioinformatics \cite{pirooznia_yang_yang_deng_2008}, credit scoring \cite{min_lee_2005}, time series forecasting \cite{cao_2003}, and weather forecasting \cite{sharma_sharma_irwin_shenoy_2011}. Text categorization or text classification is 
{to classify} documents into predefined categories. SVM and KMs work well for this task because the high dimensional text or dense concept representation can be easily mapped into a latent feature space where a linear prediction model is learned with an appropriately chosen kernel function \cite{a_2003}. 
The results of the experiments indicate that SVM with RBF kernel outperforms other classification methods \cite{joachims_1998}. The superior performance of SVM with RBF kernel in dealing with high dimensional small datasets has also been demonstrated in remote sensing \cite{fauvel_chanussot_benediktsson}, by carefully choosing feature mappings.


The performance of SVM largely depends on kernel types and it has been shown that RBF kernel support vector machine is always capable of outperforming other classifiers in various classification scenarios \cite{gumus_kilic_sertbas_ucan_2010,joachims_1998,pirooznia_yang_yang_deng_2008}. Nonetheless, in practical nonlinear modeling, SVM with standard Gaussian RBF kernel has a non-negligible limitation in separating some nonlinear decision boundaries. Thus, the analysis of RBF kernel optimization has gained much more popularity than before. The result given in \cite{amari_wu_1999} demonstrates that after introducing an information-geometric data-dependent method to modify a kernel (eg, the RBF kernel), the performance of SVM is considerably improved. Yu et al. \cite{NIPS2008_3541} enhance the kernel metrics by adding regularization into kernel machines (eg. RBF kernel SVM). 

One of the advantages of the standard SVM model is its model sparsity determined by the so-called support vectors, however the sparsity cannot be pre-determined and the support vectors have to be learned from the training data by solving a 
computationally demanding quadratic programming optimization problem \cite{HongWeiGao2018}. A massive progress in 
proposing computationally efficient algorithms for SVM models has been explored. One of the examples is the introduction of a least squares version of support vector machine (LSSVM) \cite{suykens_vandewalle_1999}. Instead of the margin constraints in the standard SVM, LSSVM introduces the equality constraints in the model formulation. The resulting quadratic programming problem can be solved by a set of linear equations \cite{suykens_vandewalle_1999}.
However, 
LSSVM is loosing of sparseness offered by the original SVM method, which leads a kernel model evaluating all possible pairs of data in the kernel function and therefore is inferior to the standard SVM model in inference for large scale data learning. To maintain the sparsity offered by the standard SVM and the equality constraints of LSSVM, researchers considered extending LSSVM for the Ramp loss function and produce sparse models with extra computational complexity, see \cite{LiuShiTianHuang2016}. This strategy has been extended to more general insensitive loss function in \cite{YeGaoShaoLiJin2019}.  Recently, Zhu et al. \cite{ZhuGaoXuYangTao2018} proposed a way to select effective patterns from training datasets for fast support vector regression learning. However, there is no extension for classification problems yet.

The need in dealing with large scale datasets motivates exploring new approaches for the sparse models under the broad framework of both SVM and KMs. Chen \cite{Chen2006} proposed a method for building a sparse kernel model by extending the so-called orthogonal least squares (OLS) algorithm \cite{ChenCowanGrant1991} and kernel techniques. It seems the OLS assisted sparse kernel model offers an efficient learning procedure particularly demonstrating good performance in nonlinear system identification. The OLS algorithm relies on a greedy sequential selection of the kernel regressors under the orthogonal requirement imposing extra computational cost.  Based on the so-called significant vectors, Gao et al. \cite{GaoShiLiu2007} proposed a more straightforward way to learn the significant regressors from training data for the kernel regression modelling.  This type of approaches has their roots in the relevance vector machine (RVM) \cite{Tipping2001}.  
RVM is implemented under the Bayesian learning
framework of kernel machine and has a comparable inference performance to the standard SVM with dramatically fewer kernel terms, offering great sparsity. 

Almost all the aforementioned modeling methods build models by learning or extracting those key data points or patterns from the entire training dataset. 
Recently, the authors proposed a new type of low rank kernel model based on the so-called simplex basis functions (SBF) \cite{HongWeiGao2018}, successfully building a sparse and fast modeling algorithm 
thus  lowering the computational cost in  LSSVM. The model size is no longer determined by the given training data while the key patterns will be learned straightaway. We further explore the idea and extend it for the so-called robust radial basis functions. 
The main contributions of this paper are summarized as follows, 
\begin{enumerate}
\item Given that the aforementioned models learn data patterns under the  the regression setting, this paper focuses on classification setting for a controlled or pre-defined model size;
\item The kernel function proposed in this paper takes the form of composition of basic basis components which are adaptive to the training data. This composition form opens the door for a fast closed form solution, avoiding the issue of kernel matrix inversion in the case of large scale datasets;
\item A new criterion is proposed for the final model selection in terms of pattern parameters of location and scale; and
\item A two-step optimization algorithm is proposed to simplify the learning procedure.
\end{enumerate}

The rest of this paper is organized as follows. In Section \ref{Sec:2}, we
present the brief background on several related models. Section \ref{Sec:3} proposes
our robust RBF kernel function and its classification model. Section \ref{Sec:4} describes the artificial
and real-world datasets and conducts several experiments to demonstrate the performance of the model and algorithm and Section \ref{Sec:5} concludes the
paper.

\section{Background and Notation}\label{Sec:2}

In this section,  we start introducing necessary notation for the purpose of presenting our model and algorithm. We mainly consider binary classification problems. For the multi-class classification problems, as usual,  the commonly used heuristic approach of ``one-vs-all'' or ``one-vs-one'' can be adopted.

Given a training dataset $\mathcal{D} = (\boldsymbol{X},\boldsymbol{t}) =  \{(\boldsymbol{x}_n, t_n)\}^N_{n=1}$ where $N$ is the number of data, $\boldsymbol{x}_n\in \mathbb{R}^D$ is the feature vector and $t_n \in \{-1, 1\}$ is the label for the $n$-th data respectively.

KM methods have been used as a universal approximator in data modeling. The core idea of the KMs is to implement a linear model in a high dimensional feature space by using a feature mapping $\boldsymbol{\phi}$ defined as \cite{SchlkopfSmola2002}
\[
\boldsymbol{x} \in \mathbb{R}^D \rightarrow \boldsymbol{\phi}(\boldsymbol{x})\in \mathcal{F},
\]
which induces a \textit{Mercer} kernel function in the input space
\[
k(\boldsymbol{x}_i, \boldsymbol{x}_j) = \langle \boldsymbol{\phi}(\boldsymbol{x}_i), \boldsymbol{\phi}(\boldsymbol{x}_j)\rangle,
\]
where $\langle \cdot, \cdot\rangle$ is the inner product on the feature space $\mathcal{F}$. 

In general, an affine linear model of KMs is defined as 
\begin{align}
y(\boldsymbol{x}) = \langle\boldsymbol{\phi}(\boldsymbol{x}), \boldsymbol{w}\rangle + b, \label{Eq:Model}
\end{align}
where $b\in\mathbb{R}$ is the bias parameter and $\boldsymbol{w}\in\mathcal{F}$ is the parameter vector of high dimensionality, most likely in infinite dimension. It is infeasible to solve for the parameter vector $\boldsymbol{w}$ directly. Instead, the so-called kernel trick transforms the infinite dimension problem to a finite dimension problem by relating the parameters $\boldsymbol{w}$ to the data as
\begin{align}
\boldsymbol{w} = \sum^N_{n=1} a_nt_n \boldsymbol{\phi}(\boldsymbol{x}_n). \label{Eq:Trick}
\end{align}
A learning algorithm will focus on solving for $N$ parameters $\boldsymbol{a} = (a_1, a_2, ..., a_N)^T\in\mathbb{R}^N$ under an appropriate learning criterion. 

For the sake of convenience, define 
\[
\boldsymbol{k}(\boldsymbol{x},\boldsymbol{X}) =\begin{bmatrix}
k(\boldsymbol{x}, \boldsymbol{x}_1) &
k(\boldsymbol{x}, \boldsymbol{x}_2)&
\cdots&
k(\boldsymbol{x}, \boldsymbol{x}_N)
\end{bmatrix}^T \in\mathbb{R}^N.
\]
Then, under \eqref{Eq:Trick}, model \eqref{Eq:Model} can be expressed in terms of new parameters $\boldsymbol{a}$ as\footnote{If we are considering a regression problem, there is no need to add $t_n$ in the model \eqref{Eq:Model_Dual2}.}
\begin{align}
y(\boldsymbol{x}) = \boldsymbol{k}(\boldsymbol{x}, \boldsymbol{X})^T (\boldsymbol{a}\circ\boldsymbol{t}) +b. \label{Eq:Model_Dual2}
\end{align}
where $\circ$ means the component-wise product of two vectors.

All the KMs algorithms are involved with the so-called kernel matrix, as defined below
\[
\boldsymbol{K} = 
\begin{bmatrix}
k(\boldsymbol{x}_1,\boldsymbol{x}_1) & \cdots & k(\boldsymbol{x}_1,\boldsymbol{x}_N) \\
 \vdots & \ddots & \vdots\\
    k(\boldsymbol{x}_N,\boldsymbol{x}_1) & \cdots & k(\boldsymbol{x}_N,\boldsymbol{x}_N)
\end{bmatrix} \in\mathbb{R}^{N\times N}
\]
and
\[
\boldsymbol{\Omega} =  \begin{bmatrix}
t_1t_1k(\boldsymbol{x}_1,\boldsymbol{x}_1) & \cdots & t_1t_Nk(\boldsymbol{x}_1,\boldsymbol{x}_N) \\
 \vdots & \ddots & \vdots\\
    t_Nt_1k(\boldsymbol{x}_N,\boldsymbol{x}_1) & \cdots & t_Nt_Nk(\boldsymbol{x}_N,\boldsymbol{x}_N)
\end{bmatrix} \in\mathbb{R}^{N\times N}.
\]
Both $\boldsymbol{K}$ and $\boldsymbol{\Omega}$ are  symmetric matrices of size $N\times N$.

In the following section, standard SVM, LSSVM and sparse least square support vector machine using simplex basis function (LSSVM-SBF) \cite{HongWeiGao2018} are outlined. 
  
\subsection{C-SVM}
The standard support vector machine (C-SVM) imposes the so-called maximal margin criterion inducing a kernel model where the parameter $\boldsymbol{a}$ (and $b$) can be obtained by solving the following dual Lagrangian problem
\begin{align}
\begin{aligned} 
    & \min_{\boldsymbol{a}}(\boldsymbol{a}\circ\boldsymbol{t})^T\boldsymbol{K} (\boldsymbol{a}\circ\boldsymbol{t})- \mathbf 1^T \boldsymbol{a},\\
    \text{s.t.}\; &\mathbf 1^T (\boldsymbol{a}\circ \boldsymbol{t})= 0,  \text{and } 0 \leq \boldsymbol{a} \leq C,
\end{aligned}\label{Eq:SVM}
\end{align}
where $\boldsymbol{1}$ is the vector with all ones in appropriate dimension. The parameter $b$ can be easily calculated from the support vectors \cite{SchlkopfSmola2002}. 

The margin criterion guarantees that the resulting kernel model \eqref{Eq:Model_Dual2} is sparse, as only those parameters $a_n$ corresponding to the support vectors $\boldsymbol{x}_n$ are non-zero. However, when $N$ is large, solving the convex quadratic programming problem  \eqref{Eq:SVM} to identify such support vectors is very time consuming.  

\subsection{LSSVM}
To reduce the computational complexity of the standard SVM, the least square support vector machine introduces the equality constraints. 

The standard LSSVM is formulated in the following programming problem  
\begin{equation}
\begin{aligned}
    & \min_{\boldsymbol{w},b,\boldsymbol{\eta}}\frac{1}{2} \|\boldsymbol{w}\|^2_{\mathcal{F}} + \frac{\gamma}{2}\sum_{n=1}^N\eta_n^2, \\
    \text{s.t. } &\;t_n(\langle\boldsymbol{w},\boldsymbol{\phi}(\boldsymbol{x_n})\rangle+b)= 1-\eta_n,\; n= 1,\cdots,N,\\
\end{aligned}\label{Eq:LSSVM}
\end{equation}
where $\gamma>0$ is a penalty parameter. 

With the given equality constraints, the Lagrangian multiplier method produces a kernel model \eqref{Eq:Model_Dual2} such that the parameters $\boldsymbol{a}$ and $b$ are given  by the following set of closed form linear equations
\begin{equation} 
  \begin{bmatrix}
    b \\
    \boldsymbol{a}
  \end{bmatrix}
  =
    \begin{bmatrix}
    0 & \boldsymbol{t}^T \\
    \boldsymbol{t} & \Omega + \boldsymbol{I}/\gamma
  \end{bmatrix}^{-1}
  \begin{bmatrix}
    0 \\
    \boldsymbol{1}
  \end{bmatrix} \label{Eq:LSSVM-Sol}
\end{equation}
where $\boldsymbol{I}$ is the identity matrix of size $N\times N$. However, the computational hurdle lies in the massive matrix inverse in \eqref{Eq:LSSVM-Sol} which has complexity of order $O(N^3)$.

\subsection{LSSVM-SBF}\label{SubSec:2.3}
Despite of a close formed solution obtained by LSSVM, the model has two main limitations. First, calculating the matrix inversion is computationally demanding and second, the model is non-sparse which means that it has to compute all possible pairs of system inputs, making the model infeasible for large-sized datasets. Alternatively, we have proposed a novel kernel method referred to as LSSVM-SBF \cite{HongWeiGao2018}, which can overcome these two issues by introducing symmetric structure in specially designed kernel function based on 
the so-called low rank Simplex Basis Function (SBF) kernel. 


The SBF $\phi_j(\boldsymbol{x};\boldsymbol\mu_j,\boldsymbol{c}_j)$ is defined as
\begin{equation}
    \phi_j(\boldsymbol{x};\boldsymbol\mu_j,\boldsymbol{c}_j)=\max\left\{0,1-\sum_{i=1}^D\mu_{i,j}|x_i-c_{i,j}|\right\}, \label{Eq:SBF}
\end{equation}
where $\boldsymbol{c}_j=[c_{1,j},\cdots, c_{D,j}]^T\in\mathbb{R}^D$ and $\boldsymbol{\mu}_j=[\mu_{1,j}, \cdots, \mu_{D,j}]^T\in\mathbb{R}^D_+$ are the center vector of the $j$th SBF function that adjusts the location and the shape vector of the $j$th SBF that adjusts the shape respectively. The proposed new kernel in \cite{HongWeiGao2018} is defined as
\begin{equation}
    k(\boldsymbol{x}',\boldsymbol{x}'')=\sum_{j=1}^M\phi_j(\boldsymbol{x}';\boldsymbol\mu_j,\boldsymbol{c}_j)^T\phi_j(\boldsymbol{x}'';\boldsymbol\mu_j,\boldsymbol{c}_j) \label{Eq:newKernel}
\end{equation}
in which the SBF kernels use only $M \ll N$ basis functions. $M$ is the pre-defined model size.  

It has been proved in \cite{HongWeiGao2018} that, under the kernel \eqref{Eq:newKernel} with the SBF \eqref{Eq:SBF}, the resulting model is  piecewise locally linear with respect to the input $\boldsymbol{x}$ as
\[
y(\boldsymbol{x}) = [\boldsymbol{\alpha}(\boldsymbol{x})]^T\boldsymbol{x} + \beta(\boldsymbol{x}).
\]
Here we have defined
\begin{equation}
\begin{aligned}
    & \beta(\boldsymbol{x})=\sum_{j\in S(\boldsymbol{x})}\theta_j(1-\sum_{i=1}^D\mu_{i,j}c_{i,j}\text{sign}(c_{i,j}-x_i))+b\\
    & \boldsymbol{\alpha}(\boldsymbol{x})=[\alpha_1(\boldsymbol{x}),\dots,\alpha_D(\boldsymbol{x})]^T, \text{in which}\\
    & \alpha_i(\boldsymbol{x})=\sum_{j\in S(\boldsymbol{x})}\theta_j\mu_{i,j}\text{sign}(c_{i,j}-x_i), i=1,\dots,D
\end{aligned}
\end{equation}
where $S(\boldsymbol{x}) \in [1, 2, ..., M]$ is the index set of $j$, satisfying condition $\sum^D_{i=1}\mu_{i,j}|x_i-c_{i,j}|<1$, and 
\[
\theta_j = \sum^N_{n=1}a_nt_n\phi_j(\boldsymbol{x}_n; \boldsymbol{\mu}_j, \boldsymbol{c}_j).
\]

With the low rank kernel structure defined as \eqref{Eq:newKernel}, the closed form solution \eqref{Eq:LSSVM-Sol} for $\boldsymbol{a}$ and $b$ can be rewritten as, see \cite{HongWeiGao2018},
\begin{align}
& \begin{bmatrix}
b \\
\boldsymbol{a}
 \end{bmatrix} = \boldsymbol{q}-\boldsymbol{P}\tilde{\boldsymbol{\Phi}}(\boldsymbol{I}+\tilde{\boldsymbol{\Phi}}^T\boldsymbol{P}\tilde{\boldsymbol{\Phi}})^{-1}\tilde{\boldsymbol{\Phi}}^T\boldsymbol{q}, \label{Eq:Sol2}
\end{align}
where
\begin{align*}
\boldsymbol{P}= 
  \frac{1}{N}
  \begin{bmatrix}
    -1/\gamma & \boldsymbol{t}^T \\
    \boldsymbol{t} & \gamma(N\boldsymbol{I}-\boldsymbol{t}\boldsymbol{t}^T)
  \end{bmatrix},\ \    
\boldsymbol{q} = 
    \frac{1}{N}
    \begin{bmatrix}
        \boldsymbol{t}^T\boldsymbol{1} \\
        \gamma(N\boldsymbol{I}-\boldsymbol{t}\boldsymbol{t}^T)
    \end{bmatrix}   
\end{align*}
and
\begin{equation*}
    \tilde{\boldsymbol{\Phi}}=
    \begin{bmatrix}
    0 & \cdots & 0 \\
    \boldsymbol{t}\circ\boldsymbol{\phi}_1 & \cdots & \boldsymbol{t}\circ\boldsymbol{\phi}_M
    \end{bmatrix},
\end{equation*}
with $\boldsymbol{\phi}_j = [\phi_j(\boldsymbol{x}_1;\boldsymbol\mu_j,\boldsymbol{c}_j), \phi_j(\boldsymbol{x}_2;\boldsymbol\mu_j,\boldsymbol{c}_j), ..., \phi_j(\boldsymbol{x}_N;\boldsymbol\mu_j,\boldsymbol{c}_j)]^T$, i.e., the vector of basis function values at the training inputs.

The new solution \eqref{Eq:Sol2} only involves the matrix inverse of size $M\times M$, which is superior to \eqref{Eq:LSSVM-Sol} where the inverse is of size $N\times N$.

\section{The Proposed Model and Its Algorithm}\label{Sec:3}

From subsection \ref{SubSec:2.3}, we have found that the special choice of low rank SBF kernel as defined in \eqref{Eq:SBF} and \eqref{Eq:newKernel} brings model efficiency. To extend the idea of using low rank kernel, in this section, we will propose a general framework for fast algorithm and validate it with several examples. 

We would like to emphasize that our idea of using low rank kernel is inspired by the original low rank kernel approximation such as Nystr\"{o}m approximation \cite{WilliamsSeeger2001}. However the standard low rank kernel methods aim to approximate a given kernel function, while our approach involves learning   (basis) functions and constructs the kernel with composite structure in order to assist fast algorithms. 

\subsection{The Low Rank Kernels and Models}
Consider $M$ learnable ``basis'' functions
\begin{align} 
\phi_j(\boldsymbol{x}; \boldsymbol{\lambda}_j):   j = 1, 2, ..., M. \label{Eq:Learnable}
\end{align}
with adaptable parameters $\boldsymbol{\lambda}_j$ ($j=1,2,...,M$). In the case of SBF in \eqref{Eq:SBF}, we have in total $2D$ parameters 
\[
\boldsymbol{\lambda}_j = \{\boldsymbol{\mu}_j, \boldsymbol{c}_j\}.
\]

As another example, we will consider the  so-called robust RBF
\begin{equation}
\phi_j(\boldsymbol{x};\boldsymbol\mu_j,\boldsymbol{c}_j)=\exp\left\{-\sum_{i=1}^D\mu_{i,j}|x_i-c_{i,j}|\right\}. \label{Eq:RobustRBF}
\end{equation}
Similar to the SBF, while $c_{i,j}$ determines the location of $\phi_j(\boldsymbol{x};\boldsymbol{\mu}_j,\boldsymbol{c}_j)$ in the $i$th dimensional direction, $\mu_{i,j}$ restricts the sharpness of  $\phi_j(\boldsymbol{x};\boldsymbol{\mu}_j,\boldsymbol{c}_j)$ in the $i$th dimension. 
In fact, the SBF \eqref{Eq:SBF} can be regarded as the first order approximation of the robust RBF in terms of $\exp\{-t\} = 1 -t +\frac1{2!}t^2 +\cdots$. We expect the robust RBF will have better modeling capability.

More generally, each learnable basis function $\phi_j(\boldsymbol{x}; \boldsymbol{\lambda}_j)$ can be a deep neural network. We will leave this for further study.

Given a set of learnable basis functions \eqref{Eq:Learnable}, define a finite dimensional feature mapping 
\[
\boldsymbol{\phi}_r: \boldsymbol{x}\in\mathbb{R}^D\rightarrow \boldsymbol{\phi}_r(\boldsymbol{x})=\begin{bmatrix}\phi_1(\boldsymbol{x}; \boldsymbol{\lambda}_1)\\
\vdots\\
\phi_M(\boldsymbol{x};\boldsymbol{\lambda}_M)]
\end{bmatrix}\in \mathcal{F}.  
\]
This feature mapping naturally induces the following learnable low rank kernel 
\begin{align}
    k(\boldsymbol{x}',\boldsymbol{x}'')=\sum_{j=1}^M\phi_j(\boldsymbol{x}';\boldsymbol{\lambda}_j)^T\phi_j(\boldsymbol{x}'';\boldsymbol{\lambda}_j). \label{Eq:learnableKernel}
\end{align}

Consider the ``linear'' model
$y(\boldsymbol{x}) = \langle \boldsymbol{w},\boldsymbol{\phi}_r(\boldsymbol{x})\rangle + b
$ and define the following low rank LSSVM (LR-LSSVM)
\begin{equation}
\begin{aligned}
    & \min_{\boldsymbol{w},b,\boldsymbol{\eta}}\frac{1}{2} \|\boldsymbol{w}\|^2_{\mathcal{F}} + \frac{\gamma}{2}\sum_{n=1}^N\eta_n^2, \\
    \text{s.t. } &\;t_n(\langle\boldsymbol{w},\boldsymbol{\phi}_r(\boldsymbol{x_n})\rangle+b)= 1-\eta_n,\; n= 1,\cdots,N.\\
\end{aligned}\label{Eq:LR-LSSVM}
\end{equation}

LR-LSSVM problems takes the same form as the standard LSSVM \eqref{Eq:LSSVM}, however our low rank kernel carries composition structure and is learnable with adaptable parameters. In the following subsections, we propose a two-steps alternative algorithm procedure to solve the LR-LSSVM.

\subsection{Solving LR-LSSVM with Fixed Feature Mappings}\label{Subsec:LR-LSSVM}
When all the feature mappings $\phi_j (j=1,2,...,M)$ are fixed, problem \eqref{Eq:LR-LSSVM} gives back to the standard LSSVM. Denote $\boldsymbol{\eta} = [\eta_1, \eta_2, ..., \eta_N]^T$ and consider the Lagrangian function
\begin{align*}
L(\boldsymbol{w},b, \boldsymbol{\eta}; \boldsymbol{a}) =& \frac{\gamma}2\|\boldsymbol{\eta}\|^2 + \frac12 \|\boldsymbol{w}\|^2\\ -\sum^N_{n=1}a_n\{t_n(&\langle\boldsymbol{w},\boldsymbol{\phi}_r(\boldsymbol{x_n})\rangle+b) -1+\eta_n\},   
\end{align*}
where $\boldsymbol{a}=[a_1, a_2, ..., a_N]^T$ are Lagrange multipliers for all the equality constraints. We now optimize out $\boldsymbol{w}$, $b$ and $\boldsymbol{\eta}$ to give
\begin{align}
\frac{\partial L}{\partial \boldsymbol{w}}=\mathbf 0 \rightarrow &\;\; \boldsymbol{w} = \boldsymbol{\Phi}^T (\boldsymbol{a}\circ\boldsymbol{t}) \label{Eq:W}\\
\frac{\partial L}{\partial b} = 0\rightarrow &\;\; \boldsymbol{t}^T\circ\boldsymbol{a} = 0 \\
\frac{\partial L}{\partial \boldsymbol{\eta}}=\mathbf 0 \rightarrow &\;\; \boldsymbol{a} = \gamma \boldsymbol{\eta}
\end{align}
where
\begin{align}
\boldsymbol{\Phi} = \begin{bmatrix}\phi_1(\boldsymbol{x}_1;\lambda_1) & \cdots &\phi_M(\boldsymbol{x}_1;\lambda_M)\\
\vdots & \cdots & \vdots\\
\phi_1(\boldsymbol{x}_N;\lambda_1) & \cdots &\phi_M(\boldsymbol{x}_N;\lambda_M)
\end{bmatrix}\in\mathbb{R}^{N\times M}. \label{Eq:PhiA}
\end{align}
Furthermore, setting the partial derivative with respect to each Lagrange multiplier gives
\begin{align}
t_n (\langle\boldsymbol{w},\boldsymbol{\phi}_r(\boldsymbol{x_n})\rangle+b) = 1 -  \eta_n;   n = 1, 2, ..., N. \label{Eq:A}
\end{align}
Taking \eqref{Eq:W} into \eqref{Eq:A} we have
\[
\boldsymbol{t}b+(\text{diag}(\boldsymbol{t})\boldsymbol{\Phi}\boldsymbol{\Phi}^T\text{diag}(\boldsymbol{t})) \boldsymbol{a} + \frac1{\gamma}\boldsymbol{a}= \mathbf 1.  
\]
After a long algebraic manipulation, the solution for the dual problem is given by
\[
\begin{bmatrix}
b\\
\boldsymbol{a}
\end{bmatrix}= 
\begin{bmatrix}
0 & \boldsymbol{t}^T\\
\boldsymbol{t} & \text{diag}(\boldsymbol{t})\boldsymbol{\Phi}\boldsymbol{\Phi}^T\text{diag}(\boldsymbol{t}) + \boldsymbol{I}/\gamma 
\end{bmatrix}^{-1}\begin{bmatrix}
0\\
\boldsymbol{1}
\end{bmatrix}.
\]
Denote $\widetilde{\boldsymbol{\Phi}}$ the $(N+1)\times M$ matrix with one row of all zeros on the top of matrix $\text{diag}(\boldsymbol{t})\boldsymbol{\Phi}$, then the solution can be expressed as
\begin{align}
\begin{bmatrix}
b\\
\boldsymbol{a}
\end{bmatrix}= \left\{
\begin{bmatrix}
0 & \boldsymbol{t}^T\\
\boldsymbol{t} & \boldsymbol{I}/\gamma 
\end{bmatrix} + \widetilde{\boldsymbol{\Phi}}\widetilde{\boldsymbol{\Phi}}^T\right\}^{-1}\begin{bmatrix}
0\\
\boldsymbol{1}
\end{bmatrix}. \label{Eq:FinalSol}
\end{align}
Applying the matrix inversion formula to \eqref{Eq:FinalSol} results in the exactly same solution as \eqref{Eq:Sol2}. Once   $\boldsymbol{a}$ and $b$ are worked out, the final model can be written as
\begin{align}
y(\boldsymbol{x}) = \langle \boldsymbol{\Phi}^T(\boldsymbol{a}\circ\boldsymbol{t}), \boldsymbol{\phi}_r(\boldsymbol{x})\rangle + b. \label{Eq:FinalModel}
\end{align}
Define 
\[
\theta_j = \sum^N_{n=1} a_nt_n \phi_j(\boldsymbol{x}_n; \boldsymbol{\lambda}_j) 
\]
which can be calculated after $\boldsymbol{a}$ is known, 
then \eqref{Eq:FinalModel} can be expressed in terms sparse form of size $M$

\begin{align}
    y(\boldsymbol{x}) = \sum^M_{j=1}\theta_j \phi_j(\boldsymbol{x}; \boldsymbol{\lambda}_j) + b. \label{Eq:modelOutput}
\end{align}

\subsection{Training Learnable Low Rank Kernels}\label{Subsec:GD}
Given $\boldsymbol{a},b$ which are solved by the closed-form solution in the first step, we estimate the kernel parameters $\boldsymbol{\lambda}_j$ 
($j=1,\dots,M$) using a gradient descent algorithm. The algorithm seeks to maximize the magnitude of model outputs, which leads to overall further distance from the model outputs to the existing decision boundary. Taking the robust RBF functions \eqref{Eq:RobustRBF} as an example, this objective function can be expressed as
\begin{equation}
    J^{(j)}(\boldsymbol{c}_j, \boldsymbol{\mu}_j)=\sum_{n=1}^N|y(\boldsymbol{x}_n)|. \label{Eq:2}
\end{equation}
Another objective function is
\begin{equation*}
    J^{(j)}(\boldsymbol{c}_j, \boldsymbol{\mu}_j)=\sum_{n=1}^Nt_ny(\boldsymbol{x}_n),  
\end{equation*}
which gives similar results as \eqref{Eq:2}. 

Denote $\text{sign}(\boldsymbol{y})=[\text{sign}(y(\boldsymbol{x}_1)),\dots,\text{sign}(y(\boldsymbol{x}_N))]^T$. Given the objective function above, we have
\begin{align}
\begin{cases}
\displaystyle\frac{\partial J^{(j)}}{\partial\mu_{i,j}}=[\text{sign}(\boldsymbol{y})]^T\frac{\partial\boldsymbol{K}}{\partial\mu_{i,j}}(\boldsymbol{a}\circ\boldsymbol{t}) & i=1,\dots,D\\
\displaystyle\frac{\partial J^{(j)}}{\partial c_{i,j}}=[\text{sign}(\boldsymbol{y})]^T\frac{\partial\boldsymbol{K}}{\partial c_{i,j}}(\boldsymbol{a}\circ\boldsymbol{t}) & i=1,\dots,D\\
\end{cases}
  \label{Eq:35}
\end{align}
in which
\begin{align}
\begin{aligned}
\frac{\partial\boldsymbol{K}}{\partial\mu_{i,j}}=(\frac{\partial}{\partial\mu_{i,j}}\boldsymbol{\phi}_j)\boldsymbol{\phi}_j^T+\boldsymbol{\phi}_j(\frac{\partial}{\partial\mu_{i,j}}\boldsymbol{\phi}_j)^T\\
\frac{\partial\boldsymbol{K}}{\partial c_{i,j}}=(\frac{\partial}{\partial c_{i,j}}\boldsymbol{\phi}_j)\boldsymbol{\phi}_j^T+\boldsymbol{\phi}_j(\frac{\partial}{\partial c_{i,j}}\boldsymbol{\phi}_j)^T\\
    \end{aligned}\label{Eq:36}
\end{align}
where
\begin{align}
\begin{aligned}
\frac{\partial}{\partial\mu_{i,j}}\boldsymbol{\phi}_j=\left[\frac{\partial\phi_j(\boldsymbol{x}_1)}{\partial\mu_{i,j}},...,\frac{\partial\phi_j(\boldsymbol{x}_N)}{\partial\mu_{i,j}}\right]^T\\
\frac{\partial}{\partial c_{i,j}}\boldsymbol{\phi}_j=\left[\frac{\partial\phi_j(\boldsymbol{x}_1)}{\partial c_{i,j}},...,\frac{\partial\phi_j(\boldsymbol{x}_N)}{\partial c_{i,j}}\right]^T
\end{aligned} 
\end{align}
which are calculated by, for $i=1, ..., D$,
\begin{equation}
\begin{split}
\frac{\partial\phi_j(\boldsymbol{x}_n)}{\partial\mu_{i,j}}=-|x_{i,n}-c_{i,j}| \phi_j(\boldsymbol{x}_n;\boldsymbol\mu_j,\boldsymbol{c}_j), 
\end{split}
\end{equation}
\begin{equation}
    \begin{split}
        \frac{\partial\phi_j(\boldsymbol{x}_n)}{\partial c_{i,j}}=\mu_{i,j}\text{sign}(x_{i,n}-c_{i,j})\phi_j(\boldsymbol{x}_n;\boldsymbol\mu_j,\boldsymbol{c}_j).
    \end{split}
\end{equation}
where $\phi_j(\boldsymbol{x};\boldsymbol\mu_j,\boldsymbol{c}_j)$ is defined in \eqref{Eq:RobustRBF}.

Meanwhile, we should also consider the positivity constraints for the shape parameters vector $\boldsymbol{\mu}_j$ and thus, we have the following constrained normalized gradient descent algorithm, which is, for $i=1,\dots,D$,
\begin{equation}
 \begin{cases}
c_{i,j}=& \displaystyle c_{i,j}+\eta \cdot \frac{\partial J^{(i)}}{\partial c_{i,j}} / \left\|\frac{\partial J^{(i)}}{\partial \boldsymbol{c}_{j}}\right\|\\
\tilde{\mu}_{i,j}=&  \displaystyle\mu_{i,j}+\eta \cdot \frac{\partial J^{(i)}}{\partial\mu_{i,j}} / \left\|\frac{\partial J^{(i)}}{\partial\boldsymbol{\mu}_{j}}\right\|\\
\mu_{i,j}=& \displaystyle \max(0,\tilde{\mu}_{i,j})
\end{cases} \label{Eq:GD1}
\end{equation}
 where $\eta>0$ is a preset learning rate. 
By applying \eqref{Eq:35} to \eqref{Eq:GD1} to all $M$ Robust RBF units while keeping $b, \boldsymbol{a}$ to their current values and other RBF units constant, we manage to update all RBF kernels.

\subsection{Initialization of Robust Radial Basis Functions}
As is shown in \eqref{Eq:modelOutput}, the model requires a preset kernel model size $M$ and a set of initial kernel parameters $\boldsymbol{\lambda}_j$,
$j=1,\dots,M$. In the case of robust RBFs, both $\boldsymbol{c}_j$ and $\boldsymbol{\mu}_j$ need to be initialized. The initialization of the center vector $\boldsymbol{c}_j$ can be obtained using a clustering algorithm. We propose a $k$-medoids algorithm here to solve for the Robust RBF centers since it is more robust to unbalanced distribution of data. It seeks to divide the data points into $M$ subsets and iteratively adjust the centers $\boldsymbol{c}_j$ of each subset $S_j$ until reaching convergence while minimizing the clustering objective objection given by
\begin{equation}
    J=\sum_{j=1}^M\sum_{\boldsymbol{x}_n\in S_j}\| \boldsymbol{x}_n-\boldsymbol{c}_j \|
\end{equation}
where the centers $\boldsymbol{c}_j$ of each subset are the members of that subset. As for the initial values of the shaping parameters $\boldsymbol{\mu}_j$, we preset $\mu_{i,j}$ as a predetermined constant for all basis functions, e.g., 1s.

\subsection{The Overall Algorithm and Its Complexity}
Algorithm \ref{Alg:1}\footnote{The algorithm can be easily adopted to any learnable kernels.} summarizes the overall procedure of LR-LSSVM using the example of robust RBF kernel. The algorithm starts with the k-medoids clustering algorithm for initialization of the robust RBF centres in Section \ref{Subsec:LR-LSSVM}, then the fast LSSVM solution is achieved and the gradient descent algorithm in Section \ref{Subsec:GD} or \ref{Subsec:GD1} are alternatively applied for a predefined number of iterations. A simple complexity analysis indicates that the overall computational complexity is $O(M^2N)$ which is dominated by the gradient descent algorithm for training learnable basis functions, scaled by the iteration number. Many examples in Section \ref{Sec:4} have shown that a minor size $M$ gives competitive model prediction performance. In this sense, the newly proposed algorithm has a complexity of $O(N)$. The lower complexity benefits from the special structure of low rank kernel functions. It should be pointed out again that the proposed framework contains the SBF model in \cite{HongWeiGao2018} as a special case, that the framework can be applied for more generic extension, for example using deep neural networks for learning kernel functions.
\begin{algorithm} 
\renewcommand{\algorithmicrequire}{\textbf{Input:}}
\renewcommand\algorithmicensure {\textbf{Output:} }
\caption{The Proposed LR-LSSVM Algorithm with robust RBF kernel}\label{Alg:1}
\begin{algorithmic}[1]
\REQUIRE Dataset $\mathcal{D}$. Model size $M$, Regularization parameter $\gamma$. Initial shape parameter $\boldsymbol{\mu}_j$. Iteration number $T$.
\ENSURE The obtained model parameters $\boldsymbol{a}$, $b$, $\boldsymbol{\lambda}_j =(\boldsymbol{c}_j, \boldsymbol{\mu}_j)$ for $j=1,2,..., M$.
\STATE Apply the k-medoids clustering algorithm to initialize $\boldsymbol{c}_j$ ($j=1, 2, ..., M$). Set all $\mu_{i,j}$ to a constant $\mu$.
\FOR{$t=1, 2, ..., T$}
\STATE Form $\boldsymbol{\Phi}$ from the dataset $\mathcal{D}$ and the current kernel parameters $\boldsymbol{\lambda}_j = (\boldsymbol{c}_j, \boldsymbol{\mu}_j)$ for $j=1,2,..., M$;

\STATE Construct $\widetilde{\boldsymbol{\Phi}}$ by adding one row of $\mathbf 0$ on the top of matrix $\text{diag}(\boldsymbol{t})\boldsymbol{\Phi}$; 

\STATE Update $b$ and $\boldsymbol{a}$ according to the closed form solution \eqref{Eq:Sol2};

\FOR{$j=1, 2, ..., M$}
\STATE Apply \eqref{Eq:35} to \eqref{Eq:GD1} to adjust $\boldsymbol{\lambda}_j =(\boldsymbol{c}_j, \boldsymbol{\mu}_j)$
\ENDFOR
\ENDFOR
\end{algorithmic}
\end{algorithm}

\subsection{The Differentiable Objective Functions}\label{Subsec:GD1}

The objective defined in \eqref{Eq:2} is non-differentiable. For the purpose of  maximizing  the  magnitude  of  model  outputs, we propose the following squared objective which is differentiable, for $j=1,2,..., M$,
\begin{equation}
    F^{(j)}(\boldsymbol{c}_j, \boldsymbol{\mu}_j)=\sum_{n=1}^Ny(\boldsymbol{x}_n)^2. \label{Eq:2aa}
\end{equation}

Then according to \eqref{Eq:FinalModel}, we can write \eqref{Eq:2aa} as
\begin{align}
&F^{(j)}(\boldsymbol{c}_j, \boldsymbol{\mu}_j)
= \sum^N_{n=1} (k(\boldsymbol{x}_n, \mathbf X)^T(\boldsymbol{a}\circ\boldsymbol{t}) + b )^2 \notag\\
=& (\boldsymbol{a}\circ\boldsymbol{t})^T\mathbf K\mathbf K(\boldsymbol{a}\circ\boldsymbol{t}) + 2b \mathbf 1^T\mathbf K (\boldsymbol{a}\circ\boldsymbol{t}) + Nb^2.
\end{align}

It is not hard to prove that
\begin{align}
\frac{\partial F^{(j)}}{\partial\mathbf K} = (\boldsymbol{a}\circ\boldsymbol{t})(\boldsymbol{a}\circ\boldsymbol{t})^T\mathbf K + \mathbf K (\boldsymbol{a}\circ\boldsymbol{t})(\boldsymbol{a}\circ\boldsymbol{t})^T + 2b \mathbf 1 (\boldsymbol{a}\circ\boldsymbol{t})^T  \label{Eq:35a}
\end{align}
and the chain rule gives
\begin{align}
\frac{\partial F^{(j)}}{\partial \nu} = \text{sum} \left(\frac{\partial F^{(j)}}{\partial\mathbf K} \circ \frac{\partial \mathbf K}{\partial \nu} \right)= \text{tr}\left(\frac{\partial F^{(j)}}{\partial\mathbf K} \frac{\partial \mathbf K}{\partial \nu}\right)\label{Eq:35b}
\end{align}
where $\text{tr}()$ means the trace of matrix, $\nu$ means either $\mu_{i,j}$ or $c_{i,j}$, and $\circ$ means the matrix element-wise product. 
Combining \eqref{Eq:35a} and \eqref{Eq:35b} gives
\begin{align}
\frac{\partial F^{(j)}}{\partial \nu} =& (\boldsymbol{a}\circ\boldsymbol{t})^T\left(\mathbf K \frac{\partial \mathbf K}{\partial \nu}+ \frac{\partial \mathbf K}{\partial \nu}\mathbf K\right)(\boldsymbol{a}\circ\boldsymbol{t}) \notag\\
&+2b (\boldsymbol{a}\circ\boldsymbol{t})^T \frac{\partial \mathbf K}{\partial \nu}\mathbf 1 \label{EQ:35c}
\end{align}
where $\frac{\partial \mathbf K}{\partial \nu}$ is the matrix given by \eqref{Eq:35} and \eqref{Eq:36}. 

\section{Experimental Studies}\label{Sec:4}

\subsection{Example 1: Synthetic Dataset}

For synthetic data set in \cite{ripley_1996}, the dimension of input space is $D=2$, and the training and test sample sets are in the size of 250 and 1000 respectively. In this example, three types of models are constructed to generate classification performance comparison by using the metric of misclassification rate. For LSSVM with Gaussian RBF kernel models, the steepness $\sigma$ is set in the range of 0.5-3, step 0.5, while shrinkage $\gamma$ is all set into 5000. For the LR-LSSVM-SBF model, the parameters are preset to $M=4, \mu=0.2, T=100, \eta=0.02, \gamma=200$. 
For our proposed LR-LSSVM-Robust RBF models with absolute value, squared and targeted objective functions, the parameters are set into $M=3, \mu=0.2, \gamma=150,  \eta=0.0008,  T=100$; $M=3,  \mu=0.2,  \gamma=20,  \eta=0.0005,  T=100$ and $M=3,  \mu=0.2,  \gamma=150, \eta=0.0008, T=100$ respectively.

\begin{figure*}[tp]
\centering
\subfigure[Synthetic data: decision boundary of gaussian SVM ($\sigma=1$)]{ \label{fig:test1}
\includegraphics[width=0.40\linewidth]{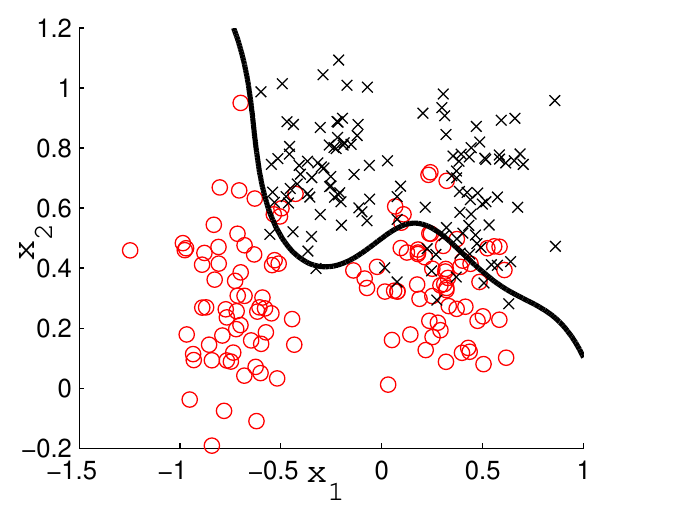} }\;\;\;\;
\subfigure[Synthetic data: decision boundary of LSSVM-SBF]{\label{fig:test2}
\includegraphics[width=0.36\linewidth]{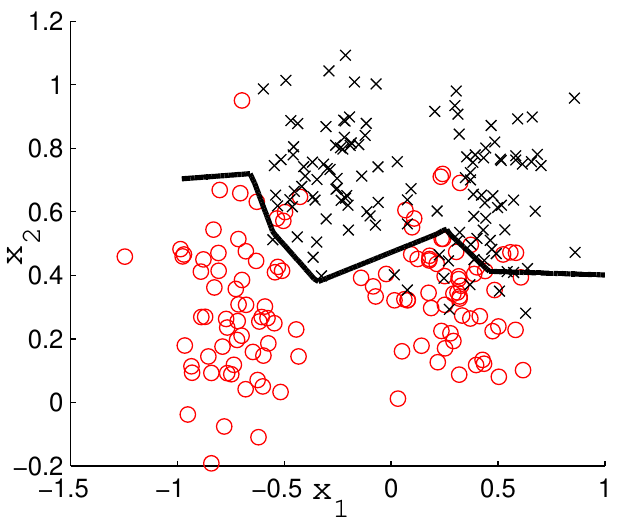} }
\subfigure[Synthetic data: decision boundary of LR-LSSVM using absolute loss function]{\label{fig:test3} 
  \includegraphics[width=0.40\linewidth]{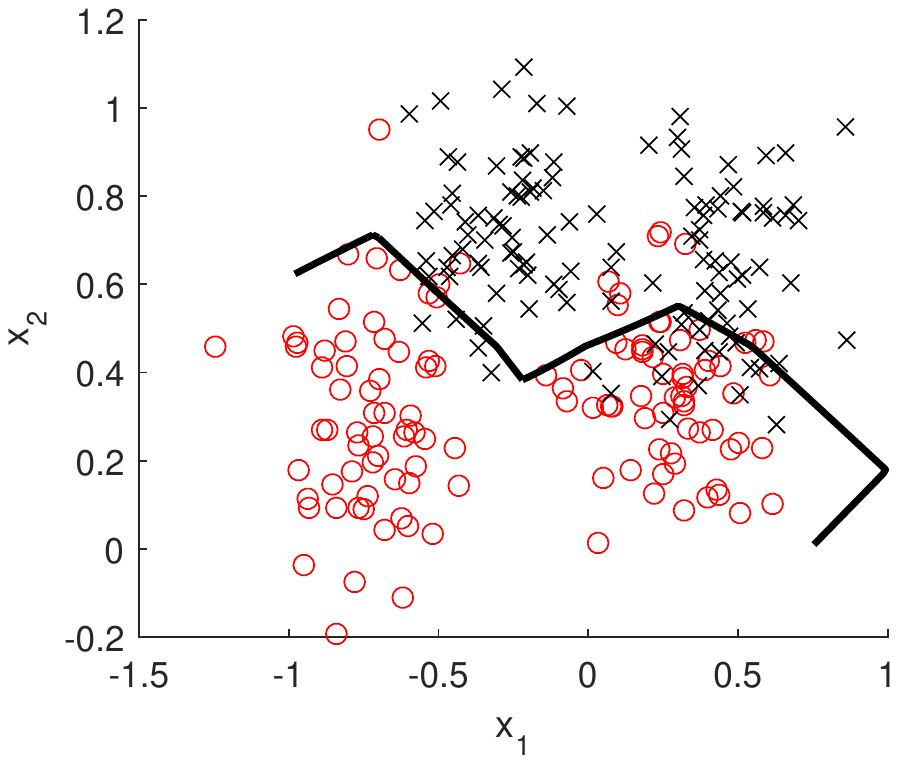}}\;\;\;\;
\subfigure[Synthetic data: decision boundary of LR-LSSVM using squared loss function]{\label{fig:test4}  
  \includegraphics[width=0.40\linewidth]{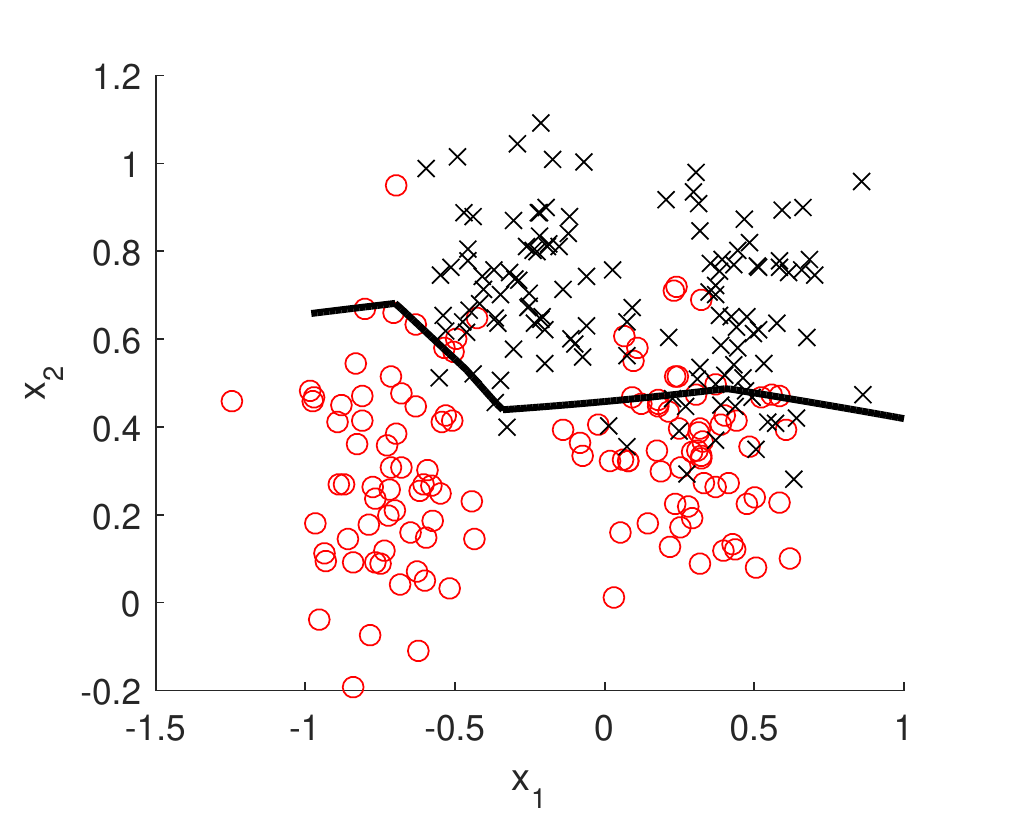}}
\caption{The experiment results for synthetic dataset}\label{Fig:1}
\end{figure*}

From the classification results shown in TABLE \ref{tab:my_label1}, we can find that the proposed LR-LSSVM-Robust RBF and LR-LSSVM-SBF models dominate all the time with the misclassification rates of around 8$\%$, while Gaussian RBF kernel models perfrom fairly poor in this case. In Fig \ref{Fig:1}, we can see that the decision boundary of LSSVM with Gaussian RBF kernel is relatively curvey and nonlinear, whereas the ones for SBF and Robust RBF are in piecewise linear forms. 

\begin{table}[ht]
\caption{The misclassification rate on synthetic data}
    \label{tab:my_label1}
    \centering
    \begin{tabular}{|>{\centering}m{3.5cm}|>{\centering}m{2.5cm}|c|}\hline
         & Testing Misclassification Rate (\%)  &Model Size\\ \hline
 LSSVM-Gaussian ($\sigma=0.5$)     & 11.40\% &  250\\ \hline
 LSSVM-Gaussian ($\sigma=1.0$)     & 9.20\% &  250\\ \hline
 LSSVM-Gaussian ($\sigma=1.5$)     & 10.40\% &  250\\ \hline
LSSVM-Gaussian ($\sigma=2.0$)     & 10.10\% &  250\\ \hline
 LSSVM-Gaussian ($\sigma=2.5$)     & 10.10\% &  250\\ \hline
LSSVM-Gaussian ($\sigma=3.0$)     & 9.80\% &  250\\ \hline
LSSVM-SBF     & 8.30\% &  4\\ \hline  
Proposed Model (abs obj.)    & 8.00\% &  3\\ \hline  
Proposed Model (square obj.)     & 8.30\% &  3\\ \hline  
Proposed Model (target obj.)     & 8.00\% &  3\\ \hline  
    \end{tabular}
\end{table}


\begin{table*}[th]
    \caption{The misclassification rate on different datasets}
    \label{table2}
\centering
\begin{tabular}{|>{\centering}m{3.5cm}|>{\centering}m{2.1cm}|c||>{\centering}m{2.1cm}|c||>{\centering}m{2.1cm}|c|}
\hline
\multirow{2}{*}{Models} & \multicolumn{2}{c||}{Titanic} & \multicolumn{2}{c||}{Diabetes} & \multicolumn{2}{c|}{German Credit}\\\cline{2-7}
&  Misclassification Rate (\%)& Mosel Size & Misclassification Rate (\%)& Model Size & Misclassification Rate (\%) & Model Size\\\hline
 RBF    & 23.3 $\pm$ 1.3 &  4 & 24.3$\pm$2.3 &  15& 24.7 $\pm$ 2.4 &  8\\ \hline
 Adaboost with RBF     & 22.6 $\pm$ 1.2 &  4& 26.5$\pm$1.9 &  15& 27.5 $\pm$ 2.5 &  8\\ \hline
 AdaBoostReg     & 22.6 $\pm$ 1.2 &  4& 23.8$\pm$1.8 &  15 & 24.3 $\pm$ 2.1 &  8\\ \hline
LPReg-AdaBoost      & 24.0 $\pm$ 4.4 &  4& 24.1$\pm$1.9 &  15& 24.8 $\pm$ 2.2 &  8\\ \hline
QPReg-AdaBoost      & 22.7 $\pm$ 1.1 &  4& 25.4$\pm$2.2 &  15& 25.3 $\pm$ 2.1 &  8\\ \hline
SVM with RBF kernel      & 22.4 $\pm$ 1.0 &  not available& 23.5$\pm$1.7 &  not available& 23.6 $\pm$ 2.1 &  not available\\ \hline
LSSVM-SBF     & 22.5 $\pm$ 0.8 &  2& 23.5$\pm$1.7 &  5& 24.9 $\pm$ 1.9 &  3\\ \hline  
Proposed Model (abs obj.)    & 22.3 $\pm$ 0.8 &  2& 23.8$\pm$1.7 &  5& 25.6 $\pm$ 2.3 &  2\\ \hline  
Proposed Model (square obj.)    & 22.6 $\pm$ 1.5 &  3& 23.5$\pm$2.0 &  4& 24.7 $\pm$ 1.9 &  2\\ \hline  
Proposed Model (target obj.)    & 22.4 $\pm$ 0.8 &  2& 24.7$\pm$2.0 &  5& 25.6 $\pm$ 2.4 &  2\\ \hline  
\end{tabular} 
\end{table*}

\subsection{Example 2: Titanic Dataset}
For the Titanic data set in \cite{Ratsch2001},  it has 100 realizations and each has 150 training samples and 2051 test samples respectively. The original data has the input dimension of $D=3$. We compare the prediction accuracy of various Adaboost-based models and the LR-LSSVM models 
over the test samples. For the LR-LSSVM-SBF model, the parameters are set into $M=2, \mu=0.2, T=100, \eta=0.05, \gamma=50000$, while for the proposed models with absolute value, squared and targeted objective functions, the parameters are set as $M=2$, $\mu=0.03$, $\gamma=50000$, $\eta=0.0005$, $T=100$; $M=3$, $\mu=0.001$, $\gamma=500000$, $\eta=0.0001$, $T=100$ and $M=2$, $\mu=0.001$, $\gamma=50000$, $\eta=0.0001$, $T=100$ respectively. 

The result of the proposed models is shown in TABLE \ref{table2} (columns 2 \& 3) together with the first six other results quoted from \cite{Ratsch2001} and the seventh result quoted from \cite{HongWeiGao2018}. Generally, LR-LSSVM-SBF and the proposed LR-LSSVM models with Robust RBF kernel outperform other models and all the LR-LSSVM models are sparse with only 2 terms (except for the model with squared loss function). 
Also, we can observe that the LR-LSSVM models with absolute value and targeted objective function have similar prediction results. Overall, the proposed models with absolute value and targeted objective functions perform the best with the lowest misclassification rate and standard deviation since the final model size of the Robust RBF kernels is only 2, which makes it easy for the models to explain the data. 

\subsection{Example 3: Diabetes Dataset}

For diabetes data set in \cite{Ratsch2001}, it has 100 groups of training and test samples individually, with the size of training set equal to 468 and the size of test set equal to 300. The input space of this example is $D=8$. Similar to the main structure of titanic data set, here, for comparison, we will use ten different models and the measurement metric of average misclassification rate as well. 
For the LR-LSSVM-SBF model, the parameters are set into $M=5, \mu=0.2, T=100, \eta=0.05, \gamma=50000$, while for the proposed models with absolute value, squared and targeted objective functions, the parameters are set as $M=5, \mu=0.01, T=100, \eta=0.001, \gamma=50000$; $M=4$, $\mu=0.001$, $\gamma=50000$, $\eta=0.001$, $T=100$ and $M=5$, $\mu=0.0001$, $\gamma=50000$, $\eta=0.001$, $T=100$ respectively.

The modeling results in TABLE \ref{table2} (columns 4 \& 5) show that the performance of the proposed LR-LSSVM-Robust RBF models with absolute value and squared objective functions are competitive in the ten models with the classification accuracy almost ranking at the top. Moreover, it can be seen that the SBF kernel and the proposed Robust RBF kernel bring sparsity into the LR-LSSVM models, which considerably increases the programming speed during computation.

\subsection{Example 4: German Credit Dataset}
Similarly, German credit dataset in \cite{Ratsch2001} has 100 realizations of training and test sets. Each realization contains 700 training samples and 300 test samples. The original data has the 20 features. We evaluate the misclassification rate of our proposed models with various objective functions and the LR-LSSVM-SBF model along with the six other models. 
For the parameters of the LR-LSSVM-SBF model, we set $M=2$, $\mu=0.005$, $\gamma=200000$, $\eta=0.003$, $T=100$ while for the proposed LR-LSSVM-Robust RBF models with absolute value, squared and targeted objective functions, the parameters are set into $M=2,  \mu=0.005,  \gamma=200000,  \eta=0.003,  T=100$ for all three cases. 

The results of the four models are listed in TABLE \ref{table2} (columns 6 \& 7) together with the first six other results quoted from \cite{Ratsch2001}. For this data set both LR-LSSVM-SBF and LR-LSSVM-Robust RBF do not perform as well as they do in the previous data sets. However, the prediction accuracy together with the standard deviation are still comparable. Additionally, it can been seen that the model size of the four models is relatively small compared to other models.

\subsection{Summary}
Overall, we can notice that the proposed squared objective model perfroms well in high dimensional datasets, which include the diabetes and german examples in our demonstration, whereas the proposed absolute value and targeted objective models are more suitable for low dimensional input, which are the synthetic and titanic datasets in our cases. Moreover, there is no relation between input dimension and chosen model size since in the four result tables, we can observe that the final selected $M$ is relatively random in general.

\section{Conclusions}\label{Sec:5}
In this paper we have generalized a widely-applied framework for fast LR-LSSVM algorithm and then extended this idea to the novel robust RBF kernel. After initialising the proposed kernel parameters with k-medoids clustering, the working procedures of training algorithm are alternating between fast least square closed form solution for $\boldsymbol{a},b$ and gradient descent for $\boldsymbol{c},\boldsymbol{\mu}$ sub-algorithms. For the gradient descent section, three criteria are offered - two non-differentiable (absolute value and targeted) and one differentiable (squared) objective functions with squared objective working better in the case of high dimensional input and the rest targeting more on low dimensional data. In the end, for the aim of demonstrating the effectiveness of our proposed algorithm, simple synthetic as well as several real-world data sets are validated in comparison with other known approaches.



\bibliographystyle{IEEEtran}
\bibliography{reference_JadeFlute} 

\end{document}